\documentclass[10pt,journal,compsoc]{IEEEtran}

\ifCLASSOPTIONcompsoc
  \usepackage[nocompress]{cite}
\else
  \usepackage{cite}
\fi

\usepackage{graphicx}
\usepackage{booktabs} 
\usepackage{amsmath}
\usepackage{hyperref}
\usepackage{xcolor}
\usepackage{multirow}
\usepackage{subcaption}
\usepackage{algorithm}
\usepackage{algpseudocode}
\usepackage{changes}
\usepackage[bottom]{footmisc}
\definechangesauthor[name={Iztok Fister}, color=red]{iztok}

\usepackage{enumitem}
\newlist{RQ}{enumerate}{1}
\setlist[RQ]{label=RQ-\arabic*:}

\usepackage{array,booktabs,ragged2e}
\newcolumntype{R}[1]{>{\RaggedRight\arraybackslash}p{#1}}
\newcolumntype{C}[1]{>{\centering\arraybackslash}p{#1}}
\newcolumntype{L}[1]{>{\RaggedLeft\arraybackslash}p{#1}}

\ifCLASSINFOpdf
\else
\fi

\begin{document}

\title{Information Cartography in Association Rule Mining}

\author{Iztok Fister Jr.,~\IEEEmembership{Member,~IEEE,}
        Iztok Fister,~\IEEEmembership{Member,~IEEE}
\IEEEcompsocitemizethanks{\IEEEcompsocthanksitem Iztok Fister Jr. and Iztok Fister are with the Faculty of Electrical Engineering and Computer Science, University of Maribor, Koro\v{s}ka cesta 46, 2000 Maribor, Slovenia. Email: iztok.fister1@um.si
\IEEEcompsocthanksitem \color{red}\textsuperscript{\textcopyright} 2021 IEEE.  Personal use of this material is permitted.  Permission from IEEE must be obtained for all other uses, in any current or future media, including reprinting/republishing this material for advertising or promotional purposes, creating new collective works, for resale or redistribution to servers or lists, or reuse of any copyrighted component of this work in other works.}}\color{black}

\markboth{IEEE Transactions on Emerging Topics in Computational Intelligence}
{Fister and Fister: Bare Demo of IEEEtran.cls for Computer Society Journals}

\IEEEtitleabstractindextext{%
\begin{abstract}
Association Rule Mining is a machine learning method for discovering the interesting relations between the attributes in a huge transaction database. Typically, algorithms for Association Rule Mining generate a huge number of association rules, from which it is hard to extract structured knowledge and present this automatically in a form that would be suitable for the user. Recently, an information cartography has been proposed for creating structured summaries of information and visualizing with methodology called ''metro maps''. This was applied to several problem domains, where pattern mining was necessary. The aim of this study is to develop a method for automatic creation of metro maps of information obtained by Association Rule Mining and, thus, spread its applicability to the other machine learning methods. Although the proposed method consists of multiple steps, its core presents metro map construction that is defined in the study as an optimization problem, which is solved using an evolutionary algorithm. Finally, this was applied to four well-known UCI Machine Learning datasets and one sport dataset. Visualizing the resulted metro maps not only justifies that this is a suitable tool for presenting structured knowledge hidden in data, but also that they can tell stories to users.
\end{abstract}

\begin{IEEEkeywords}
information cartography, metro map, Evolutionary Algorithms, Machine Learning, eXplainable Artificial Intelligence
\end{IEEEkeywords}}

\maketitle

\IEEEdisplaynontitleabstractindextext

\IEEEpeerreviewmaketitle

\IEEEraisesectionheading{\section{Introduction}\label{sec:introduction}}
\IEEEPARstart{A}{ssociation} Rule Mining (ARM) is a Machine Learning (ML) method for discovering the interesting relations between attributes in a huge transaction database. Usually, algorithms for ARM generate a huge number of association rules collected in large datasets. Hence, we are confronted with the problem of how to  extract structured knowledge from the large datasets and then present this knowledge automatically to the user in a suitable form. Nowadays, this problem is being solved with eXplainable Artificial Intelligence (XAI) methods that explain to users how decisions, usually affecting human lives, are made by deployed AI models in practice. While the models created by ML methods, like linear/logistic regression, decision tree, k-nearest neighbors, are transparent and, thus, easy to understand, the models built by methods, like tree ensembles, random forest, deep learning, are too complex and therefore act as a black-box to users~\cite{barredo2020explainable}. Consequently, danger has arisen on creating and using decisions that are not justifiable, legitimate, or do not allow detailed explanations of their behavior~\cite{gunning2017explainable}.

To overcome this problem, the XAI helps users to understand, trust, and effectively manage the emerging generation of AI partners~\cite{gunning2019xai}. Such explainable systems already help users to understand the information generated from the ML models in medicine, transportation, security and economy, among others~\cite{barredo2020explainable}. These systems use several explanation techniques, like: explanation by simplification, feature relevance explanation, explanation by example and visual explanation~\cite{barredo2020explainable}. Recently, a metro map of the information concept has been developed~\cite{shahaf2012trains} that is capable of creating structured summaries of information. The name was selected as a metaphor for a real cartographic map, i.e. in the same way that these maps help people understand their surroundings, metro maps help them understand the information landscape~\cite{shahaf2013information}. Moreover, visualization with metro maps (also information cartography) can tell stories to users, on the one hand, and provide them with good directions, on the other. Indeed, the metro map consists of a set of lines, where each line interprets the same story from a different aspect. Metro stops on these lines introduce salient pieces of information, while the interrelations among these pieces ensure the plot of the story. 

The first algorithm for ARM was Apriori, which was proposed back in 1994 by Agrawal~\cite{agrawal1994fast}. Apriori is still the most popular algorithm for mining association rules. This was also considered to be one of the top 10 algorithms in ML~\cite{wu2008top}. Some of the other well-established algorithms in this domain are Eclat~\cite{zaki2000scalable}, FP-Growth~\cite{han2000mining}, Genetic Association Rules (GAR)~\cite{mata2002discovering}, the Multi-Objective Differential Evolution algorithm formining Numeric Association Rules (MODENAR)~\cite{alatas2008modenar}, and BatMiner~\cite{fister2019batminer}. 

Several algorithms were proposed for reducing the redundant or even meaningless association rules. For instance, Feng et al.~\cite{feng2016soft} employed an algorithm for rule mining based on the theory of soft sets. Recently, the logical formulas over soft sets were proposed by Feng et al.~\cite{feng2020maximal} for solving the same problem. On the other hand, Luna et al.~\cite{luna2018apriori} used a series of algorithms based on the MapReduce framework and applied this for mining frequent patterns hidden in big data. To reduce long execution times by nature-inspired population-based algorithms for ARM, Djenouri et al.~\cite{djenouri2018mining} developed a Genetic Algorithm running on Clusters of GPUs (CGPUGA) that improved performance of state-of-the-art significant for ARM on big datasets. \cite{djenouri2014pruning,djenouri2017combining}

Interestingly, the visualization of association rules has rarely been treated in literature. Indeed, the papers which referred to this topic can be summarized in the following review: The authors in~\cite{wong1999visualizing} presented a design that is able to handle hundreds of multiple antecedent association rules in a three dimensional display with minimum human interaction, low occlusion percentage, and no screen swapping. The authors in~\cite{hofmann2000visualizing} show that the use of Mosaic plots and their variant, called Double Decker plots, can be applied for visualizing association rules. Ong et al.~\cite{ong2002crystalclear} prototyped the two visualizations, called grid view and tree view, for visualizing the association rules in their application called CrystalClear. Appice and Buono~\cite{appice2005analyzing} presented a graph-based visualization that supports data miners in the analysis of multi-level spatial association rules, while Herawan et al.~\cite{herawan2009smarviz} proposed an approach for visualizing soft maximal association rules. Hahsler~\cite{hahsler2017arules} proposed the R-extension package arulesViz for visualizing the association rules using the most popular visualization techniques. A very interesting interactive visualization technique, which lets the user navigate through a hierarchy of association rule groups, is presented by Hahsler and Chelluboina~\cite{hahsler2011visualizing}. The authors in paper~\cite{jiang2008finite} explored Hasse diagrams for the visualization of Boolean association rules. Fister et al.~\cite{Fister2019Discovering} have proposed a method for identifying dependencies among mined association rules based on population-based metaheuristics and complex networks. However, there are also generic tools, like the CloseViz~\cite{carmichael2010closeviz} and the SPMF open-source data mining library Version 2~\cite{fournier2016spmf}, specialized primarily in pattern mining, offering visual implementation of discovered data mined by ML algorithms that could also be used for visualization of ARM.

Information cartography is a process of building models (i.e., information maps) based on non-geographical data~\cite{old2002information}. It draws on the metaphor of an information landscape populated by information landscapes forming an information map, and enables an analysis of data having a ''geographic hook''. In order to be visualized, information needs to be spatialized, i.e., transformed into a coordinate system and placed on a surface such as a map. The first definition of the term in the sense of visualization was made by Stephen Paling in~\cite{paling2000information}, where he established that maps are among the best information systems and require little documentation to be used commonly and to be understandable. 

The majority of people are linear thinkers and process information in a linear manner, i.e., analytically. Linear thinking is known as a step-by-step progression, where a response to one step must be elicited before another step is taken~\cite{castillo2014thinking}. It is connected with the logical, analytical, and sequential processing of information in contrast to circular, non-linear, and parallel ones. Typically, this kind of thinker operates in a two-dimensional world where time is of the utmost importance~\cite{horner2008extraordinary}. The concept of information maps in ARM comply strongly with the perception of linear thinkers, where attributes in metro lines describe a linear sequence of attributes, while the mutual connections between metro lines reveal how the attributes in one metro lines affect the attributes in others, and vice versa. These linear sequences can also be seen from a time point of view. This type of information cannot be obtained by the other ARM visualization methods (e.g., scatter plot, mosaic plot, complex networks etc.).

A seminal work in automatic creation of the metro maps was performed by Shahaf et al. in 2012~\cite{shahaf2012trains} that treated their construction as an optimization problem. The proposed method was applied to analyzing the concise structured set of documents, from which structured summaries of information were created. In paper~\cite{shahaf2012metro}, the same authors solved the problem of information overload arisen in scientific literature by proposing the same methodology. The zoomable metro maps, proposed by Shahar et al.~\cite{shahaf2013information}, help users that might be interested in information of different levels of granularity. In paper~\cite{shahaf2015metro}, Shahaf et al. showed that metro maps can even tell stories, as well as provide good directions. Interestingly, the authors of the majority of the reviewed papers proposed deterministic methods for constructing the metro maps.

The aim of the study is to develop a method for creating the metro maps of ARM information automatically. The proposed method acts as follows: The ARM information are hidden in ARM datasets produced by algorithms for ARM in the form of implication rules $X\Rightarrow Y$. In general, these rules are represented as conjunctions of more antecedent and more consequent attributes. At first, each complex rule is simplified to a set of simple ones consisting of one antecedent and one consequent. These rules serve as building blocks for the construction of an attribute graph. The attribute graph consists of nodes representing attributes (e.g. $X$, $Y$), and direct edges denoting an implication relation (e.g. $X\Rightarrow Y$). The construction of a metro map is defined as an optimization problem that searches for the best metro lines within the attribute graph. The problem is solved using an Evolutionary Algorithm (EA), because it can be applied to the problems where no specific knowledge is discovered and enables parameters to be set adaptively. Finally, the metro map is visualized. 

The main contributions of this paper can be summarized as follows:
\begin{itemize}
    \item the first application of information cartography for visualizing the association rules,
    \item widening the applicability of information cartography to ARM,
    \item supplementing the set of visual explanation techniques for post-hoc explainability with information cartography. 
\end{itemize}
The software is planned to be included into the Universal ARM Solver package~\cite{fedora32uarmsolver}.

The paper is structured as follows: Section~\ref{definition} introduces the basic information needed for understanding the subject that follows. In Section~\ref{sec:method}, the proposed method for information cartography in ARM is illustrated in detail. Section~\ref{results} presents the results of the method by constructing metro maps of information obtained by five well known datasets, while the paper concludes with a summary of the performed work and outlines directions for future work.

\section{Basic notation}
\label{definition}
The present section consists of two subsections. The former introduces the problem of ARM, while the latter describes the formal definition of the objectives necessary for identifying the information cartography in ARM as an optimization problem.

\subsection{Association Rule Mining}
ARM can be defined formally as follows: Let us assume, a set of objects $O=\{o_1,\ldots,o_N\}$ and transaction dataset $T_D=\{T\}$ are given, where each transaction $T$ is a subset of objects $T\subseteq O$. Then, an association rule is defined as an implication:
\begin{equation}
    X\Rightarrow Y,
\end{equation}
where $X\subset O$, $Y\subset O$, and $X\cap Y=\emptyset$. In order to estimate the quality of the mined association rule, two measures are defined: confidence and support. The confidence is defined as:
\begin{equation}
    \mathit{conf}(X\Rightarrow Y)=\frac{n(X\cup Y)}{n(X)},
\end{equation}
whereas support is:
\begin{equation}
    \mathit{supp}(X\Rightarrow Y)=\frac{n(X\cup Y)}{|T|},
\end{equation}
where the function $n(.)$ calculates the number of repetitions of a particular rule within $T_D$, and $|T|$ is the total number of transactions in $T_D$. Let us emphasize that two additional variables are defined, i.e. minimum confidence $C_{min}$ and minimum support $S_{min}$. These variables denote a threshold value limiting the particular association rule with lower confidence and support from being taken into consideration.

In our study, transaction databases are represented as a matrix of columns corresponding to features and rows corresponding to transactions with assigned attributes. Before using the databases, objects need to be created. Actually, the objects are represented as $<$feature,attribute$>$ pairs obtained by enumerating the feature and corresponding attributes from a domain of values that the definite feature can capture. Typically, objects in the ARM databases are designated fully by concatenation of the feature and attribute using the underlined character. 

\subsection{Formal definition of objectives} \label{sec:2.2}
The concept of a metro map is applied in order to visualize the archive of mined association rules~\cite{shahaf2013information}. In our study, the metro map is defined formally as $\mathcal{M}=(G,\Pi)$, where $G=(A,E)$ denotes an attribute graph of vertices $A=\{X_1,\ldots,X_N\}$, representing objects (i.e., $<$feature,attribute$>$ pairs), and edges $E=\{r_1,\ldots,r_M\}$, representing simple rules, together with incident function $\psi_G$ that associates an ordered pair $\psi_G(r_k)=(X,Y)$ with direct edge $r_k$, when a simple association rule exists in the form of $X\Rightarrow Y$, and $\Pi$ represents a set of metro lines $\pi\in\Pi$, where each metro line is defined as a permutation of edges $r_1,\ldots,r_{\tau}$. In the definitions, variables $N$ and $M$ denote the maximum number of vertices and maximum number of edges, respectively. Thus, the simple association rule consists of only one antecedent and one consequent, where the former is mapped to the source node $X\in G$ and the latter to the sink node $Y\in G$ of the corresponding attribute graph, while the path $X\Rightarrow Y$ leads from the source to the sink node.

In general, the association rules in the archive consist of more antecedents and more consequences, in other words:
\begin{equation}
    X_1 \wedge X_2 \wedge \ldots \wedge X_p \Rightarrow Y_1 \wedge Y_2 \wedge \ldots \wedge Y_q.
\end{equation}
The simple association rules are obtained from the mined rules by pairing each antecedent with each consequent, in other words:
\begin{equation}
    (X_1\Rightarrow Y_1),(X_1\Rightarrow Y_2),\ldots,(X_p\Rightarrow Y_q).
\end{equation}
In this process of simplifying rules, the $p\times q$ pairs of simple rules are obtained representing direct edges in the association graph.

 In the ARM domain, there are three types of attributes, i.e. categorical, numerical, and mixed. The categorical attributes consist of a domain of discrete values, while the numerical ones are assigned to a domain of continuous real values that must be discretized before use. The last attribute type can employ elements from both of the aforementioned domains. 

The objects can appear in mined association rules as: (1) antecedent only, (2) consequent only, or (3) antecedent in one and consequent in the other rule. In line with this, these are divided into three subsets, i.e. $\mathit{Source}(G)$, $\mathit{Sink}(G)$, and $\mathit{Intern}(G)$. In graph $G$, the objects in antecedent subset $X\in \mathit{Source}(G)$ represent source nodes with indegree zero, the objects in consequent subset $Y\in \mathit{Sink}(A)$ are sink nodes with outdegree zero, while the objects in the mixed subset $\langle X|Y\rangle\in \mathit{Intern}(G)$ denote the intern nodes with indegree and outdegree higher than zero. Indeed, the antecedent set consists of nodes suitable for starting metro stops on the metro lines, the consequent set for the final metro stops, while the intern set determines the intermediate metro stops and outlines a definite path towards achieving a certain end destination.

The algorithm for constructing the metro map for visualizing the association rules needs to fulfill the following four objectives:
\begin{itemize}
    \item maximum line coherence,
    \item maximum map size,
    \item high coverage,
    \item high structure quality.
\end{itemize}
The maximum line coherence limits the number of intermediate metro stops in some metro line, and is expressed by the following relation:
\begin{equation}
    \mathit{coherence}(\mathcal{M})\leq \tau,
    \label{eq:cons_1}
\end{equation}
where the variable $\tau$ determines the maximum number of intermediate metro stops. The maximum map size is referring to the number of metro lines $K$, in other words: 
\begin{equation}
    |\mathcal{M}|\leq K.
    \label{eq:cons_2}
\end{equation}
Indeed, we are interested in covering our information domain by the number of the metro lines that are close to $K$, and all the metro lines must be as coherent in the number of metro stops as possible.

The coverage estimates how well the selected metro line exploits the attributes in the transaction database. In line with this, the lift measure of association rule $\mathit{Lift}(X\Rightarrow Y)$ is used that is expressed as:
\begin{equation}
    \mathit{Lift}(X\Rightarrow Y)=\frac{\mathit{supp}(X\cup Y)}{\mathit{supp}(X)\times\mathit{supp}(Y)}.    
\end{equation}
Actually, we are interested for rules with lift value $>1$, that determines the degree to which the probability of occurrence of the antecedent and this of the consequent are dependent on one another, and makes those rules potentially useful for prediction. The coverage of the whole metro line $\pi \in \Pi$ is expressed as:
\begin{equation}
    \mathit{coverage}(\pi)=\frac{1}{|\pi|}\sum_{r\in\pi}{\mathit{Lift}(r)},
\end{equation}
where $r$ represents the particular simple association rule $X\Rightarrow Y$. Finally, the coverage of the metro map is a simple average of all the proposed metro lines, in other words:
\begin{equation}
    \mathit{coverage}(\Pi)=\frac{\sqrt{|\Pi|}}{|\mathcal{M}|}\sum_{\pi\in\Pi}{\mathit{coverage}(\pi)},
    \label{eq:cov}
\end{equation}
where $|\Pi|$ is the sum of edges (i.e., rules) in all the metro lines. Indeed, the ratio $\frac{\sqrt{|\Pi|}}{|\mathcal{M}|}$ expresses the connectivity of the observed metro line. The ratio prefers the metro line with many long metro lines, and violates those with many short ones.

The metro map structure quality demands that all the metro stops in each metro line cannot follow to the some metro stops that appear before the observed one in the other metro line. In the other words, all consequent attributes of an attribute $x_{i,k}$ in the $i$-th metro line cannot appear as an antecedent of the same attribute in all the other metro lines. For instance, let us assume two metro lines $\pi_i$ and $\pi_j$:
\begin{equation}
\begin{aligned}
    \pi_i:(x_{i,1}\Rightarrow\ldots\Rightarrow x_{i,k}\Rightarrow\underbrace{x_{i,k+1}\Rightarrow\ldots\Rightarrow x_{i,\tau_i}}_{\mathit{Cons}(x_{i,k}|\pi_i)},\\
    \pi_j:(\underbrace{x_{j,1}\Rightarrow\ldots\Rightarrow x_{j,l-1}}_{\mathit{Ante}(x_{i,k}|\pi_j)}\Rightarrow x_{j,l}\equiv x_{i,k}\Rightarrow\ldots\Rightarrow x_{i,\tau_i}
\end{aligned}
\end{equation}
Then, a violation between two metro lines $\chi(\pi_i|\pi_j)$ appears, when an intersection of the observed consequent $\mathit{Cons}(x_{i,k}|\pi_i)$ and antecedent $\mathit{Ante}(x_{i,k}|\pi_j)$ is not empty. Mathematically, the violation is expressed as:
\begin{equation}
\small
    \chi(\pi_i|\pi_j)=\begin{cases}
    0,& \text{if}~\mathit{Cons}(x_{i,k}|\pi_i)\cap\mathit{Ante}(x_{i,k}|\pi_j)=\emptyset,\\
    -1,& \text{otherwise},
    \end{cases}
\end{equation}
for $i=1,\ldots,|\mathcal{M}|-1$ and $j=i+1,\ldots,|\mathcal{M}|$.   

In the sense of the defined objectives, the problem of metro map construction is defined formally as:
\begin{equation}
    \max \mathit{coverage}(\Pi),
\end{equation}
subject to
\begin{equation}
    \mathit{sQuality}(\Pi)=\sum^{|M|-1}_{i=1}\sum^{|M|}_{j=i+1} \chi(\pi_i|\pi_j),
    \label{eq:sQual}
\end{equation}
where $\pi_i,\pi_j\in\Pi$. Let us mention that the value of $\mathit{sQuality}(\Pi)$ in Eq.~(\ref{eq:sQual}) refers to the number of violations which occurred in constructing the metro map. When the value is zero, the valid metro map is obtained. 

\section{Proposed method}
\label{sec:method}

\begin{figure*}[!htb]
    \centering
    \includegraphics[width=0.9\linewidth]{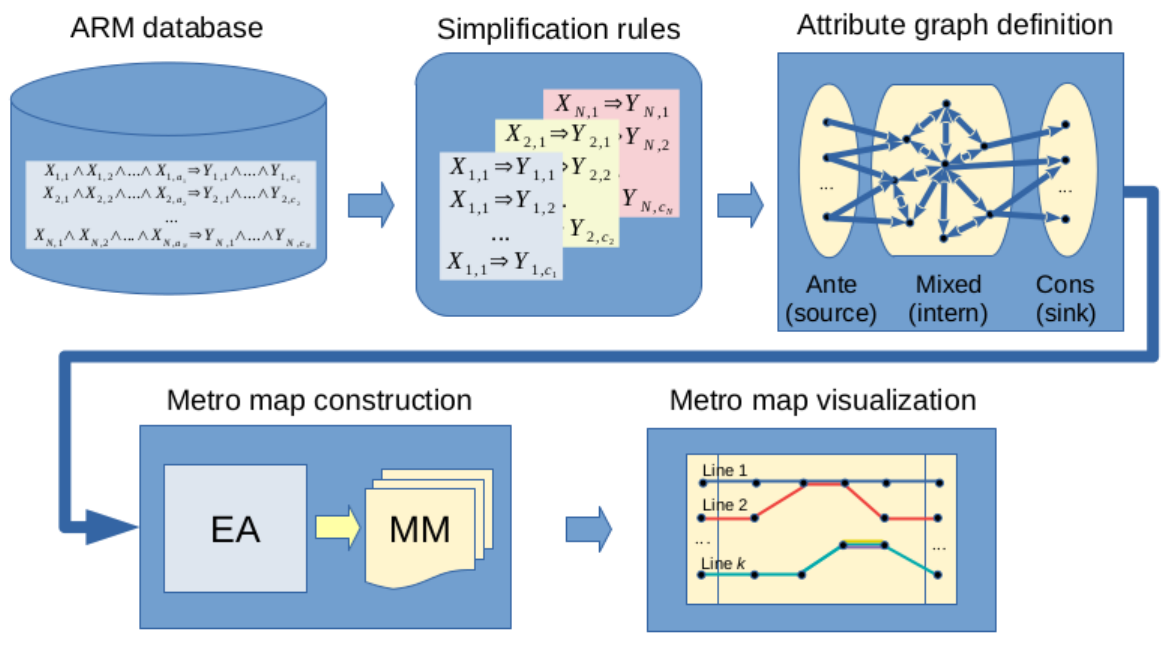}
    \caption{Concept of metro map creation in association rule mining.}
    \label{fig:MM}
\end{figure*}

The proposed method for information cartography in ARM consists of the following four steps (Fig.~\ref{fig:MM}):
\begin{itemize}
    \item creating the ARM database,
    \item association rule simplification,
    \item attribute graph definition,
    \item metro map construction,
    \item metro map visualization.
\end{itemize}
The ARM database is a result of the ARM algorithm, where the modern stochastic population-based nature-inspired algorithms, like BatMiner, can also be used instead of the classical approaches, like Apriori or FPGrowth. 

On the other hand, the results of the ARM algorithms are slightly confusing for information cartography in the sense that all objects in the mined association rules can emerge as antecedent in one and consequent in another rule. 
In a mathematical sense, we must ensure that any of the following three inequations: $\mathit{Source}(G)\neq \emptyset$, $\mathit{Sink}(G)\neq \emptyset$, and $\mathit{Intern}(G)\neq \emptyset$ is valid. The aforementioned conditions are satisfied in our study by proper filtering, where the set of the most similar rules is searched for by using the linear programming algorithm.

The main characteristics of the ARM databases are that the mined association rules are in a broad form with many antecedent, as well as consequent, attributes. This form is not appropriate for the creation of attribute graphs and, therefore, rules need to be transformed into a simplified form. The simplification rule procedure was discussed in detail in the previous section and therefore avoided here. 

The third step is dedicated for creating the attribute graph $G=(A,E)$, where all the simplified association rules are incorporated in the adjacent matrix $\mathbf{A}_G=(a_{i,j})$ of dimension $N\times N$, where: 
\begin{equation}
    a_{i,j}=\left \{ \begin{matrix}
    1, & \text{if}~X\Rightarrow Y,\\
    0, & \text{otherwise}.
    \end{matrix} \right.
\end{equation}
It should be mentioned that no loops are allowed in this graph because of $a(i,i)=0$ for $i=1,\ldots,N$. In order to characterize the type of attribute graph, a metric Average Path Length (APL) is introduced that identifies the end-to-end hop distances among all possible node pairs over the graph~\cite{manoj2018complex}. Finally, the classification of attributes in the three distinguished sets (i.e. $\mathit{Source}(G)$, $\mathit{Intern}(G)$, and $\mathit{Sink}(G)$) is also performed in this step.

In the remainder of the section, the fourth step is described, where the metro map $\mathcal{M}=(G,\Pi)$ is constructed, while the visualization of the metro map is examined in the next section.

\subsection{EA for metro map construction}
Although several authors of the existing methods for creating metro maps proposed deterministic algorithms, the EA was introduced in our study. The motivation behind using this kind of stochastic nature-inspired population-based algorithms was due to their advantages over traditional optimization algorithms (such as gradient-based methods) that can be summarized as follows~\cite{fogel1997advantages}: (1) Conceptual simplicity, (2) Broad applicability, (3) Their ability to outperform traditional methods on real problems, (4) The potential to use knowledge and hybridize with the other methods, (5) Parallelism, (6) Their robustness to dynamic changes, (7) Their capability of self-optimization, and (8) Their ability to solve problems that have no known solutions.

The EA for metro map construction demands modifications of the following algorithm components~\cite{eiben2015introduction}:
\begin{itemize}
    \item representation of a solution,
    \item variation operators (i.e. crossover and mutation),
    \item survivor and parent selection,
    \item fitness function evaluation,
    \item initialization,
    \item termination condition.
\end{itemize}
The problem of metro map construction is constrained, because every feasible solution must satisfy the relation $\mathit{sQuality}(\Pi)=0$. 

\subsubsection{Representation of solutions} \label{sec:repr}
Each solution in the population of $\mathit{Np}$ individuals represents a metro map $\mathbf{y}_i$ that is encoded as follows:
\small
\begin{equation}
    \mathbf{y}_i=(\underbrace{n_{i}}_{\text{Control variable}},\underbrace{\mathbf{x}_{i,1},\mathbf{x}_{i,2}\ldots,\mathbf{x}_{i,n_i}}_{\text{Description of metro map $\Pi$}}),
\end{equation}
\normalsize
where the first part of the representation is dedicated for control meta-parameter $n_i$ that determines the number of metro lines and determines individuals of variable length. The second part consists of detailed descriptions for particular metro lines, expressed as:
\begin{equation}
   \mathbf{x}_{i,j}=\{x_{i,j,1},\ldots,x_{i,j,|\mathbf{x}_{i,j}|}\},\quad \text{for}~i=1,\ldots,n_1, 
\end{equation}
where each element $x_{i,j,k}$ for $k=1,\ldots,|\mathbf{x}_{i,j}|$ encodes a specific simple association rule $X\Rightarrow Y$, and $|\mathbf{x}_{i,j}|$ determines the number of associative rules within the metro line $\mathbf{x}_{i,j}\in\mathbf{y}_{i}$. The elements are ordered into a sequence of implication rules: 
\begin{equation}
    x_{i,j,1}\Rightarrow x_{i,j,2}\Rightarrow x_{i,j,|\mathbf{x}_{i,j}|-1}\Rightarrow x_{i,j,|\mathbf{x}_{i,j}|}.
    \label{eq:implic}
\end{equation}
in such a way that each consequence of the $k$-th rule appears as an antecedent in the $k+1$-th rule, in other words: If consecutive elements $x_{i,j,k}$ and $x_{i,j,k+1}$ encode the rules $X_l\Rightarrow Y_l$ and $X_{l+1}\Rightarrow Y_{l+1}$ in the metro line $j$, then the relation $Y_l=X_{l+1}$ must hold.

Indeed, the representation of solutions also includes control parameter $n_{i}$ besides the problem variables. This means that the parameter controlling the maximum number of metro lines undergoes acting the variation operators and thus adapts to the fitness landscape online.

\subsubsection{Variation operators}
The proposed algorithms support two variation operators, i.e. crossover and mutation. While the crossover operates on metro lines as a whole, the mutation is also capable of modifying the structure inside a particular metro line. Obviously, the application of crossover is controlled using the probability of crossover $p_c$, and mutation using the probability of mutation $p_m$. 

\paragraph{\textbf{Crossover}} is defined as follows: At first, for each target metro map $\mathbf{y}^{(\mathit{org})}$, a trial metro map $\mathbf{y}^{(\mathit{tri})}$ is created using the same control parameters. Then, a parent metro map $\mathbf{y}^{(\mathit{par})}$ is selected randomly and the metro lines for the trial are taken either from the target or parent metro maps according to probability $p_c$.  Mathematically, this crossover is expressed as:
\small
\begin{equation}
\mathbf{x}^{(\mathit{tri})}_{i,j}=\begin{cases}
\begin{cases}
\mathbf{x}^{(\mathit{par})}_{i,j}, & \text{if}~|\mathbf{y}^{(\mathit{par})}_{i}|< |\mathbf{y}^{(\mathit{org})}_{i}|,\\ 
\emptyset, & \text{otherwise},\\
\end{cases} & \text{if }\text{U}(0,1)\leq p_c, \\
\mathbf{x}^{(\mathit{org})}_{i,j}, & \text{otherwise},\\
\end{cases}
\end{equation}
\normalsize
for $i=1,\ldots,|\mathbf{y}^{(\mathit{org})}|$. Consequently, the operator can produce infeasible solutions in two cases:
\begin{itemize}
    \item the size of the parent metro map is smaller than the size of the trial, i.e. $|y{(\mathit{par})}_{i,j}|<|x{(\mathit{par})}_{i,j}|$,
    \item the first antecedent of the metro line representing the starting metro stop is replicated twice.
\end{itemize}
In the first case, the corresponding metro line is deleted from the trial solution, while in the second, the solution from the parent solution is preferred.

\paragraph{\textbf{Mutation}} modifies the structure of the metro line as follows: At first, the position of mutation $k$ is selected randomly according to the probability of mutation $p_m$. Then, an antecedent $X_k$ is extracted from a corresponding association rule $\langle X_k\Rightarrow Y_k\rangle$, and the new consequent $Y^{'}_k$ is attached to the rule among all the possible consequent as defined by the graph $G=(A,E)$. Finally, the remaining path from $r^{'}_{k+1}$ towards the drain is generated randomly.

Mathematically, the mutation is presented as follows: Let us assume the metro line $\mathbf{x}_{i,j}=(x_{i,j,k})$ for $k=1,\ldots,|\textbf{x}_{i,j}|$ is given, where each element represents an association rule $x_{i,j,k}=X_l\Rightarrow Y_l$, and a position of mutation $k\in[1,|\textbf{x}_{i,j}|]$ is selected according to the probability of mutation $p_m$, in other words:
\tiny
\begin{equation*}
\mathbf{x}_{i,j}=(\underbrace{\langle X_1\Rightarrow Y_1\rangle}_{x_{i,j,1}},...,\underbrace{\langle X_k\Rightarrow Y_k\rangle}_{x_{i,j,k}},\underbrace{\langle X_{k+1}\Rightarrow Y_{k+1}\rangle}_{x_{i,j,k+1}},...,\underbrace{\langle X_{L}\Rightarrow Y_{L}\rangle}_{x_{i,j,L}}).
\end{equation*}
\normalsize
The result of the mutation is expressed as follows:
\tiny
\begin{equation*}
\mathbf{x}^{'}_{i,j}=(\underbrace{\langle X_1\Rightarrow Y_1\rangle}_{x_{i,j,1}},...,\underbrace{\langle X_k\Rightarrow Y_k^{'}\rangle}_{x^{'}_{i,j,k}},\underbrace{\langle X^{'}_{k+1}\Rightarrow Y^{'}_{k+1}\rangle}_{x^{'}_{i,j,k+1}},...,\underbrace{\langle X^{'}_{L}\Rightarrow Y^{'}_{L}\rangle}_{x^{'}_{i,j,L}}).
\end{equation*}
\normalsize
where $L=|\textbf{x}_{i,j}|$ and all the modified values of the metro line are denoted by apostrophes. Let us mention that the operation also has an impact on the metro line length $L$ that can be increased or decreased within the allowed maximum metro line length (i.e. $L\leq \tau$).

\subsubsection{Survivor and parent selection}
One-to-one selection is applied as an operator of survivor selection that is borrowed from Differential Evolution (DE)~\cite{storn1997differential}. This selection works on the whole metro map $\mathbf{y}_i$. Mathematically, it is expressed as follows:
\begin{equation}
\label{eq:de_sel}
 \mathbf{x}^{\mathit{(org)}}_{i}=\begin{cases}
          \mathbf{y}^{\mathit{(tri)}}_{i}, &\text{if } f(\mathbf{y}^{\mathit{(tri)}}_{i} \leq f(\mathbf{y}^{\mathit{(org)}}_{i}), \\
		  \mathbf{y}^{\mathit{(org)}}_{i}, &\text{otherwise}\,,
        \end{cases}
\end{equation}
where the better between trial $\mathbf{y}^{(tri)}_i$ and target $\mathbf{y}^{(org)}_i$ vector is preserved as the candidate solution map that proceeds into the next generation.

Parent selection is applied by a crossover operator for generating the trial solution. Although there are many parent selection operators, the parent selection implemented in our study selects the parent randomly among the other population members.

\subsubsection{Fitness function evaluation}
The problem of constructing metro maps is constrained in its nature. However, constraints can be handled in several ways in EC~\cite{eiben2015introduction}. In our case, penalizing the infeasible solutions is employed within the proposed EA, where the value of a penalty function is assigned to the fitness function, when its value is less than zero. When the penalty function becomes zero, the coverage function takes the initiative. Mathematically, the fitness function is expressed as follows:
\begin{equation}
    f(\mathbf{y}_i)=\begin{cases}
    \mathit{coverage}(\mathbf{y}_{i}), & \text{if}~\mathit{sQuality}(\mathbf{y}_{i})=0,\\
    \mathit{-sQuality}(\mathbf{y}_{i}), & \text{otherwise}.
    \end{cases}
    \label{eq:fit}
\end{equation}
The task of the optimization algorithm is to maximize the value of the fitness function. The design of such the fitness function is capable of preferring the more promising simple association rules on the one hand, and ensures that the constructed metro lines are proper.  Let us mention that the proposed fitness function also ensures the good behavior of the algorithm on highly non-separable datasets, where the metro-lines are highly interrelated. 

\subsubsection{Initialization}
Initialization of individuals in the population is performed randomly. At first, the number of metro lines is generated randomly from the interval $[2,K]$. Then, the unique source node is selected from the set of source nodes. For each source node, the random path in the attribute graph $G=(A,E)$ is searched for so that the maximum metro line length $\tau$ is not exceeded.  

\subsubsection{Termination condition}
In our study, the EA is terminated, when the fitness improvement is zero for more than the predefined $\mathit{Threshold}$ number of generations. The termination condition parameter is adaptive, because it demands a feedback from the evolutionary search process in order to determine how big the loss of the variability is in improving an average of the fitness value within the population of solutions.

\subsubsection{Pseudo-code of the algorithm}
The pseudo-code of the proposed EA for constructing metro maps is illustrated in Algorithm~\ref{alg:evol}, in which all components of a general EA as proposed by Eiben and Smith in~\cite{eiben2015introduction} can be indicated. Actually, all these components are described in detail in the previous subsections. Therefore, here, we are focused on a description of the common principles used by solving the construction problem.

\begin{algorithm}[htb]
\caption{Pseudo-code of an EA.}
\label{alg:evol}
\begin{algorithmic}[1]
\Procedure{evolutionary\_algorithm}{}
\State INITIALIZE\_population\_randomly;
\State EVALUATE\_each\_candidate\_solution;
\While {TERMINATION\_CONDITION\_not\_met}
\State SELECT\_PARENTS;
\State RECOMBINE\_pairs\_of\_parents;
\State MUTATE\_resulting\_offspring;
\State EVALUATE\_each\_candidate\_solution;
\State SELECT\_SURVIVOR\_solutions;
\EndWhile
\State \textbf{return} best\_solution
\EndProcedure
\end{algorithmic}
\end{algorithm}

Indeed, the construction process must fulfill four objectives, where maximum coherence and maximum map size are obeyed implicitly by the variation operators (functions 'RECOMBINE\_pairs\_of\_parents' and 'MUTATE\_resulting\_offspring'). This means that only the feasible solutions are taken into consideration, while the others are deleted from the population. Which parents enter into the variation process is determined by the parent selection operator (function 'SELECT\_PARENTS'). The other two objectives (i.e., high coverage and high structure quality) are captured in the fitness function of the evolutionary process (function 'EVALUATE\_each\_candidate\_solution') that operates on the 'generate-and-test' principle. In this sense, only the best solutions according to the fitness function can survive and transfer their good characteristics into the next generations (function 'SELECT\_SURVIVOR\_solutions'). In order to run the EA, also initialization (function 'INITIALIZE\_population\_randomly'), termination condition (function 'TERMINATION\_CONDITION\_not\_met'), and representation of solution (see Section~\ref{sec:repr}) are incorporated within the proposed EA.

The remainder of the section is devoted to analysis of the time complexity of the proposed algorithm as well as to discussion of the parameters controlling the behavior of particular EA components.

\paragraph{\textbf{Complexity and running time}} Time complexity depends on the problem size (the population size) and the search space dimension (the length of individuals)~\cite{beyer2014convergence}. However, EAs are stochastic. This means that there are no guarantees for the algorithm to reach an optimum. Moreover, this condition might never get satisfied and, consequently, such algorithm could never stop. As a result, performance measures, like the maximum number of generations or evaluations of the fitness function, or population diversity drop under some threshold, serve for approximating the optimum of a given fitness function~\cite{eiben2015introduction}. Consequently, we are interested about the mathematical measures in the EA community (i.e., convergence velocity, convergence rate, convergence order), describing the convergence behavior of EAs more than the time complexity. In line with this, the better the convergence rate, the better the EA.

\paragraph{\textbf{Parameters}} 
 Setting parameters has a crucial impact on the performance of the algorithms. Although an algorithm for metro map construction depends on even six parameters, we can propose some hints for their proper tuning. For instance, the maximum number of metro lines is a self-adaptive control parameter, where the proper value is drawn by the EA from some predefined interval of integer values. Thus, the reasonable minimum value of the interval should be set to 2, while the maximum value to the maximum number of the starting metro stops (i.e., $\mathit{Source}(G)$) found in the attribute graph. The last value can be delivered easily by the proposed method. The termination condition in our study is an adaptive control parameter that also could be set loosely depending on the precision of the calculation. The parameter population size needs to enable bias between exploration and exploitation firmly within the evolutionary search process, while setting the parameters probability of crossover and probability of mutation complies with the theory of Evolutionary Computation~\cite{eiben2015introduction}. Finally, setting the optimal number of metro stops stays the only control parameter that needs explicit tuning, in which help of users is also welcome by estimating the quality of the produced metro maps. However, the reasonable interval of values for this parameter should be set as follows: the minimum value could be set to 3, while the maximum value to the average path length as found in the corresponding attribute graph.

\section{Results and discussion}
\label{results}
The purpose of our experimental work was to show that the huge amount of data obtained by algorithms for ARM can be extracted automatically in the form of structured knowledge called information cartography in ARM and visualized by metro maps of ARM information. These maps can help users understand information in many knowledge domains. In line with this, we applied our proposed method to five different datasets that accompany data from various domains, e.g. biology, chemistry, and sports. 

In summary, using metro maps has several advantages for users. In line with this, the experimental study includes the following research questions:
\begin{RQ}
    \item extract the most important association rules from a huge amount of data,
    \item find the more $<$feature,attribute$>$ pairs by automated extracting of information and to direct the attention of users to them,
    \item apply the proposed visualization method to new application domain,
    \item derive the story that metro maps narrate.
\end{RQ}

The information cartography in ARM is a very complex task that comprises of five tasks as discussed in Section~\ref{sec:method}. Therefore, the results of each phase (except association rule simplification) are illustrated in detail in the remainder of the paper. Especially, we focus on exposing the characteristics of the algorithm for constructing metro maps.

\subsection{ARM datasets creation}
In our study, evaluating the efficiency of the proposed method for information cartography in ARM was performed on four public datasets from the UCI Machine Learning repository~\cite{dua2019uci} and one Sport dataset, consisting of real data obtained from device trackers worn by sport athletes (i.e. cyclists) during their sport training sessions~\cite{Fister2019Discovering}. 

The characteristics of the aforementioned datasets are illustrated in Table~\ref{tab:exp_tab1}. Let us emphasize that the results of ARM for the first four datasets were obtained using the Apriori algorithm~\cite{agrawal1994fast}, while for the Sport dataset they were obtained by using the BatMiner~\cite{fister2019batminer}. 
\begin{table}[!htb]
\begin{center}
\small
\caption{The ARM datasets characteristics.}
\label{tab:exp_tab1}
\begin{tabular}{ l|r|r|r|r|r }
\hline
\multirow{2}{*}{Dataset} & \multicolumn{3}{c|}{Characteristics} & \multicolumn{2}{c}{Mined rules} \\ \cline{2-6}
 & Instan. & Featur. & Attrib. & All & Filtered  \\	\hline
Mushroom & 8,124 & 23 & 126 & 24,408 & 998 \\
Iris & 150 & 5 & 24 & 182 & 88 \\
Abalone & 4,177 & 9 & 28 & 36,388 & 2,779 \\
Wine & 178 & 13 & 55 & 5,483 & 2,355 \\
Sport & 80 & 14 & 87 & 4,191 & 10 \\
\hline
\end{tabular}
\normalsize
\end{center}
\end{table}

Indeed, the table consists of two parts, in which the first indicates the characteristics of particular datasets, like the number of instances, the number of features, and the number of attributes, while the second contains the results of the algorithm for ARM. The results are presented in two columns: The column 'All' denotes the number of all mined association rules, while the column 'Filtered' includes the number of all rules obtained after filtering. 

In a nutshell, the characteristics of the aforementioned datasets are as follows: The Mushroom dataset contains logical rules for mushrooms, indicating if a specific one is poisonous or edible. The dataset includes descriptions of hypothetical samples corresponding to 23 species of gilled mushrooms, where each species is identified as definitely edible, definitely poisonous, or not recommended. The last class was combined with the poisonous one. As a result, there are two final classes, i.e. the edible (included into 865 rules as a consequent), and the poisonous mushrooms (included into 133 rules as a consequent). All samples are discrete values.

The Iris dataset collects transactions describing characteristics for the classification of three different iris plants (i.e. Setosa, Versicolour, and Virginica), where each transaction is constructed from five features with 21 numerical attributes. Thus, these transactions are divided equally among three classes (33.3~\% for each of the three classes).

The Abalone dataset is devoted to predicting the age of abalone from physical measurements~\cite{dua2019uci}. All measurements are numerical. The age of abalone is determined by cutting the shell through the cone, staining them, and then counting the number of rings through a microscope. Other measurements, which are easier to obtain, are used to predict the age. There are four classes originally included in the dataset. Thus, the first one is included into 69.95~\% of rules as a consequent, the second one in 30.05~\% of rules as a consequent, while the third one is not included in any association rule as a consequent and is, therefore, eliminated from the observation.

The data in the Wine dataset are obtained as the results of a chemical analysis of wines grown in the same region in Italy, but derived from three different cultivars. The analysis determined the quantities of 13 constituents (features) found in each of the three types of wines: the first is included into 59 rules as a consequent, the second included in 71  rules as a consequent, and the third included in 48  rules as a consequent. In summary, the number of attributes is equal to 19, where each of these is numerical. 

A Sport dataset was produced from the TCX files of a professional, male cyclist with many years of experience, who donated his data voluntary for the purposes of this study. Every training session was tracked by a wearable sports watch and the data imported into a training dataset. In addition to these performance data, data for estimation of the athlete’s psycho-physical conditions were also merged with each training session by the sports trainer. As a result, the measurements are of mixed types, i.e., numerical and discrete. Let us emphasize that this dataset does not include any results of classification.  

\subsection{Attribute graph definition}
In this step, the rules in the filtered ARM dataset are transformed into simple rules, at first. Because this step is trivial, it is omitted from a detailed discussion here. Obviously, the simple rules serve as building blocks for attribute graph definition. Actually, the results of the attribute graph definition are presented in Table~\ref{tab:exp_tab2} 
\begin{table}[!htb]
\begin{center}
\small
\caption{Characteristics of created attribute graphs.}
\label{tab:exp_tab2}
\begin{tabular}{ l|r|r|r|r|r|r }
\hline
\multirow{2}{*}{Dataset} & \multicolumn{3}{c|}{Graph} & \multicolumn{3}{c}{Story important nodes} \\	\cline{2-7}
 & Node & Edge & APL & Source & Intern & Sink \\	\hline
Mushroom & 16 & 116 & 23 & 4 & 10 & 2 \\
Iris & 18 & 38 & 5 & 10 & 5 & 3 \\
Abalone & 22 & 124 & 8 & 8 & 12 & 2 \\
Wine & 37 & 308 & 11 & 11 & 23 & 3 \\
Sport & 21 & 52 & 6 & 5 & 7 & 9 \\
\hline
\end{tabular}
\normalsize
\end{center}
\end{table}
which, in addition to the attribute graph specification for the particular dataset (i.e., number of nodes, edges, and APL), also depicts the sizes of the antecedent, consequent, and mixed sets that actually represent the number of source, sink, and intern nodes in the attribute graph, respectively. Together, these nodes are also called story important nodes.

As can be seen from the table, the Iris and Sport datasets are relatively simple from the problem solving point of view due to only five intern nodes (i.e. attributes). This restricts the algorithm for a metro map construction to search for solutions in a smaller search space, especially when we assume that the majority of the other nodes in the corresponding attribute graph are identified as source and sink ones. The remaining attribute graphs are more complex due to the higher number of nodes, as well as edges.

\subsection{Metro map construction}
The EA for the metro map construction used parameters set as illustrated in Table~\ref{tab:param}, from which it can be seen that the population size $\mathit{Np}=100$ was applied in all experiments. Actually, this value ensures a good bias between exploration and exploitation, and avoids the evolutionary search process getting stuck in local optima. Also, both probabilities, $p_c$ and $p_m$ were set according to the guidelines valid in the EC community. The EA was terminated when the EA did not not improve the fitness values during the last 100 generations (parameter $\mathit{Threshold}$). Obviously, the number of metro lines $K$ is problem dependent, while its value refers to the number of source nodes in the attribute graph. Therefore, the optimal value of this parameter was self-adapted in our study. The most specific was setting of the length of metro lines $\tau$ that depends on the average path length of the constructed attribute graph. This parameter is also problem specific on the one hand and strong affects the quality of the constructed metro maps on the other. Because the quality of metro maps is left to the judgment of the users, the proper setting of this parameter needs to be made by a user manually, by selecting the best solution from a collection of constructed metro maps. Each dataset was optimized 25 times, and the results of the best run obtained by constructing the metro maps based on five aforementioned datasets were selected for further analysis. This means that the run with the highest fitness of the 25 repeats was observed. 
\begin{table}[!htb]
\begin{center}
\small
\caption{Parameter setting of EA for metro map construction.}
\label{tab:param}
\begin{tabular}{ l|c|r }
\hline
Parameter & Abbreviation & Value \\	\hline
Population size & $\mathit{Np}$ & 100 \\
Probability of crossover & $p_c$ & 0.5 \\
Probability of mutation & $p_m$ & 0.01 \\
Period of fitness stagnation & $\mathit{Threshold}$ & 100 \\
Number of metro lines & $K$ & $[2,|\textit{Source}(G)|\footnotemark[1]]$ \\
Optimum metro line length & $\tau$ & $[3,\text{APL}_G\footnotemark[2]]$ \\
\hline
\end{tabular}
\normalsize
\end{center}
\end{table}
\footnotetext[1]{Number of source nodes found in attribute graph.}
\footnotetext[2]{APL of the corresponding attribute length.}

In the remainder of the section, the EA for constructing the metro maps is analyzed in detail. The analysis was conducted on the Wine dataset that is complex enough to reveal the characteristics of the algorithm. Obviously, a similar analysis could be performed on the other datasets as well.

\subsubsection{Analysis of the algorithm}
The results of the EA for constructing the metro maps obtained on the Wine dataset by varying the metro line lengths $\tau\in[3,11]$ are illustrated in Table~\ref{tab:result}, where the metro maps are presented with metro lines of path lengths 3, 5, 8, and 11. A legend, transforming the object numbers within the association rules to $<$feature,attribute$>$ pairs, is attached to the table. Interestingly, the metro map with metro line length of $\tau=5$ revealed the best result according to the fitness function. 

\begin{table*}[hbt]
\centering
\begin{minipage}{\linewidth}

\begin{center}
\caption{The results of EA for metro map construction obtained by the Wine dataset.}
\label{tab:result}
\begin{tabular}{ c|l|l|l|l }
\hline
ML & $\tau=3$ & $\tau=5$ (Best) & $\tau=8$ & $\tau=11$ \\	\hline
1 & 31$\Rightarrow$30$\Rightarrow$7 & 26$\Rightarrow$1$\Rightarrow$4$\Rightarrow$30$\Rightarrow$7 & 31$\Rightarrow$30$\Rightarrow$0$\Rightarrow$15$\Rightarrow$23$\Rightarrow$24$\Rightarrow$21$\Rightarrow$14 & 32$\Rightarrow$9$\Rightarrow$0$\Rightarrow$6$\Rightarrow$23$\Rightarrow$22$\Rightarrow$24$\Rightarrow$19$\Rightarrow$10$\Rightarrow$12$\Rightarrow$14 \\
2 & 32$\Rightarrow$9$\Rightarrow$7 & 17$\Rightarrow$2$\Rightarrow$4$\Rightarrow$3$\Rightarrow$7 & 32$\Rightarrow$9$\Rightarrow$27$\Rightarrow$23$\Rightarrow$24$\Rightarrow$19$\Rightarrow$21$\Rightarrow$14 & 31$\Rightarrow$30$\Rightarrow$0$\Rightarrow$21$\Rightarrow$6$\Rightarrow$23$\Rightarrow$24$\Rightarrow$1$\Rightarrow$27$\Rightarrow$11$\Rightarrow$25 \\
3 & 17$\Rightarrow$3$\Rightarrow$7 & 34$\Rightarrow$8$\Rightarrow$4$\Rightarrow$30$\Rightarrow$7 & 35$\Rightarrow$22$\Rightarrow$23$\Rightarrow$1$\Rightarrow$24$\Rightarrow$19$\Rightarrow$21$\Rightarrow$14 & 34$\Rightarrow$8$\Rightarrow$3$\Rightarrow$4$\Rightarrow$30$\Rightarrow$0$\Rightarrow$21$\Rightarrow$23$\Rightarrow$24$\Rightarrow$19$\Rightarrow$25 \\
4 & 33$\Rightarrow$22$\Rightarrow$14 & 35$\Rightarrow$22$\Rightarrow$24$\Rightarrow$21$\Rightarrow$14 & 34$\Rightarrow$8$\Rightarrow$4$\Rightarrow$30$\Rightarrow$0$\Rightarrow$21$\Rightarrow$6$\Rightarrow$25 &  \\
5 & 29$\Rightarrow$24$\Rightarrow$14 & 31$\Rightarrow$30$\Rightarrow$0$\Rightarrow$21$\Rightarrow$25 & 28$\Rightarrow$8$\Rightarrow$10$\Rightarrow$5$\Rightarrow$2$\Rightarrow$3$\Rightarrow$23$\Rightarrow$25 &  \\
6 & 26$\Rightarrow$22$\Rightarrow$14 & 28$\Rightarrow$4$\Rightarrow$30$\Rightarrow$0$\Rightarrow$25 &  \\
7 & 35$\Rightarrow$22$\Rightarrow$14 & 32$\Rightarrow$9$\Rightarrow$27$\Rightarrow$23$\Rightarrow$25 &  \\
8 & 34$\Rightarrow$8$\Rightarrow$25 & 33$\Rightarrow$22$\Rightarrow$24$\Rightarrow$21$\Rightarrow$25 &  \\
9 & 20$\Rightarrow$11$\Rightarrow$25 &  &  \\
\hline
\end{tabular}
\end{center}
\end{minipage}

\begin{minipage}{\linewidth}
\vspace{+.5cm}
\end{minipage}

\begin{minipage}{\linewidth}
\centering

\begin{center}
\small
\label{tab:legend}
\begin{tabular}{ |c|l|c|l|c|l| }
\hline
Nr. & Attribute & Nr. & Attribute & Nr. & Attribute \\	\hline
0 & alcalinity\_(15.45-20.3] & 12 & color\_intensity\_(-inf-4.21] &  26 & color\_intensity\_(7.14-10.07] \\
1 & ash\_(2.295-2.7625] & 14 & class\_2 &  27 & nonflavanoid\_phenols\_(0.395-0.5275] \\
2 & malic\_acid\_(-inf-2.005] & 17 & hue\_(1.095-1.4025] &  28 & nonflavanoid\_phenols\_(-inf-0.2625] \\
3 & alcohol\_(12.93-13.88] & 19 & magnesium\_(93-116] &  29 & malic\_acid\_(2.005-3.27] \\
4 & flavanoids\_(2.71-3.895] & 20 & total\_phenols\_(1.705-2.43] &  30 & proanthocyanins(1.995-2.7875] \\
6 & proanthocyanins(1.2025-1.995] & 21 & total\_phenols\_(-inf-1.705] &  31 & alcohol\_(13.88-inf) \\
7 & class\_1 & 22 & flavanoids\_(-inf-1.525] &  32 & alcohol\_(-inf-11.98] \\
8 & color\_intensity\_(4.21-7.14] & 23 & hue\_(-inf-0.7875] &  33 & malic\_acid\_(3.27-4.535] \\
9 & flavanoids\_(1.525-2.71] & 24 & proanthocyanins(-inf-1.2025] &  34 & magnesium\_(116-139] \\
10 & hue\_(0.7875-1.095] & 25 & class\_3 &  35 & nonflavanoid\_phenols\_(0.5275-inf) \\
11 & magnesium\_(-inf-93] & &   &   &  \\
\hline
\end{tabular}
\normalsize
\end{center}
\end{minipage}
\end{table*}

As can be seen from the table, each metro lines obeys the maximum number of metro stops strictly. For instance, all metro lines in the corresponding metro map in column 'Path=3' consist of three metro stops (i.e., a start, one intern, and a destination metro stop). This is also valid for the all the other path lengths. 

In summary, the number of metro lines decreases with increasing the metro stops. Moreover, some sequences of metro stops in metro lines of lower path lengths form sub-sequences that can be detected in metro lines of higher path lengths. For instance, the sub-sequence '$34\Rightarrow8$' is detected in ML-8 of $\tau=3$, in the '$34\Rightarrow8\Rightarrow4\Rightarrow30$' of ML-3 by $\tau=4$, and in the '$34\Rightarrow8\Rightarrow4\Rightarrow39\Rightarrow0\Rightarrow21$' of ML-4 by $\tau=4$. Interestingly, this sub-sequence proliferates into two sub-sequences in ML-3 of $\tau=11$, i.e., '$34\Rightarrow8$' and '$4\Rightarrow39\Rightarrow0\Rightarrow21$' connected by the metro stop 3. These sub-sequences present building blocks that connect themselves into more promising metro lines of higher path length. On the other hand, some more expressed destination metro stops may disappear from the metro maps. For instance, there exist metro maps with metro lines of higher path lengths (e.g., $\tau=8$ and $\tau=11$) with no destination metro stop of 'class\_1'.

In order to identify the characteristics of the proposed EA for constructing the metro maps, two experiments were conducted further, in which we analyze:
\begin{itemize}
\item the influence of the adaptive population size on the results,
\item the influence of the metro line length.
\end{itemize}
The purpose of the first test was to determine how big was the error caused by using the adaptive termination condition, while the second one how the metro line length $\tau$ affected the quality of the constructed metro maps. Let us also mention that the analysis of Wine dataset was employed in this test.

\paragraph{\textbf{Influence of the adaptive population size}} In this test, the population size was varied in the interval $\mathit{Np}\in\{50,100,200,500,1000\}$. Thus, the five instances of the results were obtained, to which the results produced by the EA using the adaptive termination condition were also added. In the last case, the number of generations needed is unknown in advance. However, the results represent fitness function values obtained by the corresponding instance. The values were collected according to different metro line length, where this parameter was varied in the interval $\tau\in[3,11]$ in a step of one. 

The results of the test are illustrated analytically in Table~\ref{tab:anal}, in which the best fitness function values are presented in bold.
\begin{table}[!htb]
\begin{center}
\small
\caption{Influence of the generation number analytically.}
\label{tab:anal}
\begin{tabular}{ c|rrrrrr }
\hline
$\tau$ & \multicolumn{1}{c}{50} & \multicolumn{1}{c}{100} & \multicolumn{1}{c}{200} & \multicolumn{1}{c}{500} & \multicolumn{1}{c}{1,000} & Adapt. \\ \hline
3	&	27.31	&	27.34	&	27.70	&	28.31	&	28.31	&	28.31	\\
4	&	32.21	&	32.70	&	33.08	&	33.19	&	33.19	&	33.19	\\
5	&	\textbf{34.10}	&	34.43	&	35.13	&	35.76	&	\textbf{35.76}	&	\textbf{35.76}	\\
6	&	31.95	&	33.39	&	\textbf{35.44}	&	\textbf{35.53}	&	35.56	&	35.53	\\
7	&	31.90	&	33.63	&	33.63	&	34.41	&	34.91	&	34.41	\\
8	&	31.28	&	32.08	&	32.69	&	33.94	&	34.02	&	33.94	\\
9	&	28.15	&	28.84	&	29.89	&	30.40	&	30.41	&	30.34	\\
10	&	29.64	&	30.04	&	30.04	&	30.43	&	30.45	&	30.04	\\
11	&	29.36	&	29.89	&	29.94	&	29.94	&	30.19	&	29.94	\\
\hline
\end{tabular}
\normalsize
\end{center}
\end{table}
 It can be seen from the table that the highest fitness function values are obtained by the metro line lengths $\tau=5$ and $\tau=6$. Mainly, the values below and above these metro line lengths are lower.

The same results according to different number of metro stops $\tau\in[3,11]$ are depicted graphically in Fig.~\ref{fig:main}a, where the dependence of the fitness values are presented according to the generations. Additionally, the observed generations at $G=\{50,100,200,500,1000\}$ are denoted by reference lines in the graph. 
\begin{figure*}[htb]
\centering
\begin{minipage}{.49\linewidth}
  \includegraphics[width=\linewidth]{path.pdf}
  \caption{subfigure}{Influence of the generation number graphically.}
  \label{img1}
\end{minipage}
\begin{minipage}{.49\linewidth}
  \includegraphics[width=\linewidth]{adapt.pdf}
  \caption{subfigure}{Influence of the adaptive termination condition.}
  \label{img2}
\end{minipage}
\caption{The results as affected by the number of generations.}
\label{fig:main}
\end{figure*}

As can be seen from the graph, the EA converges to the optimal values very fast, i.e., only 200 generations was enough for the convergence for all path lengths. There is less improvement detected by increasing the number of generations from 500 to 1000.

The last finding is considered in detail by analysis of the error in fitness function caused by introducing the adaptive population size. In line with this, the results of the EA using adaptive population size were compared with the results obtained by the EA with the fixed population size of higher value (e.g., $G=1000$). The results of this test are presented in Table~\ref{tab:err}, which consists of columns denoting fitness values obtained by both algorithms, the number of generation necessary for obtaining the results, and the difference in fitness values between both algorithms. 
\begin{table}[!htb]
\begin{center}
\small
\caption{Error by introducing the adaptive population size.}
\label{tab:err}
\begin{tabular}{ c|rrrrr }
\hline
Path & $f_{1000}$ & $f_{\mathit{ada}}$ & $G_{1000}$ & $G_{\mathit{ada}}$ & $\Delta_f$\\ \hline
3 & 28.31 & 28.31 & 1,000 & 451 & 0.00 \\
4 & 33.19 & 33.19 & 1,000 & 707 & 0.00 \\
5 & 35.76 & 35.76 & 1,000 & 713 & 0.00 \\
6 & 35.56 & 35.53 & 1,000 & 440 & 0.02 \\
7 & 34.91 & 34.41 & 1,000 & 509 & 0.50 \\
8 & 34.02 & 33.94 & 1,000 & 455 & 0.08 \\
9 & 30.41 & 30.34 & 1,000 & 363 & 0.07 \\
10 & 30.45 & 30.04 & 1,000 & 265 & 0.41 \\
11 & 30.19 & 29.94 & 1,000 & 237 & 0.26 \\ \hline
$\sum$ & 292.82 & 291.46 & 9,000 & 4,140 & 1.35 \\
\hline
\end{tabular}
\normalsize
\end{center}
\end{table}
As can be seen from the table, the EA with adaptive population size needed for even 54~\% less generations for achieving the results that are 0.5~\% worse than those obtained by the EA using the fixed population size $G=1,000$.

In graphical form, the last results can be seen in Fig.~\ref{fig:main}b. This figure shows that the EA needs more than 500 generation by constructing the metro maps with metro lines of length $\tau=4$, $\tau=5$, and $\tau=6$ only. All the other instances needed a lower number of generations. On the other hand, these instances also contributed the highest values to the total error, and demanded an increase in the threshold value to $\mathit{Threshold}=100$.

\paragraph{\textbf{Influence of the metro path length.}}
The aim of this test was to discover how the metro line length affected the results of the EA for constructing the metro maps. In line with this, the path length of the metro lines are varied in the interval $[3,11]$ in steps of one.

Fig.~\ref{fig:main2}a illustrates the influence of the metro line lengths on the fitness values obtained by the proposed EA. The results obtained by the EA using the generation numbers changing in interval $G\in\{50,100,200,500,1000\}$ are compared with the results obtained by its counterpart using the adaptive population size.
\begin{figure*}[htb]
\centering
\begin{minipage}{.49\linewidth}
  \includegraphics[width=\linewidth]{fit.pdf}
  \caption{subfigure}{Influence of the metro line length.}
  \label{img3}
\end{minipage}
\begin{minipage}{.49\linewidth}
  \includegraphics[width=\linewidth]{nml.pdf}
  \caption{subfigure}{Influence of the number of metro lines.}
  \label{img4}
\end{minipage}
\caption{Searching for the characteristics of the optimal metro lines.}
\label{fig:main2}
\end{figure*}

As can be seen from the figure, the results of the EA using the adaptive generation size is comparable with the results of the same using the fixed generation size of $G=1000$. Interestingly, the best results according to the fitness value were obtained by the metro lines of length $\tau=4$, while the instances below and above this values were worse. Obviously, this behavior was caused due to a connectivity factor, expressed as a ratio of the total number of edges per metro line.

The last test was devoted to discovering how the number of metro lines evolves during the typical evolutionary cycle. In line with this, the number of metro lines in the best solution according to the fitness value was measured at the beginning (i.e., at $G=0$) and at the end (i.e., $G=1,000$) of the evolutionary run. 

The results of the EA for constructing the metro maps are depicted in Fig.~\ref{fig:main2}b, from which it can be seen that the number of metro lines decreases when the path length of the metro lines was lower, and increased in the opposite case. 

\subsection{Metro map visualization}
In this subsection, the best results generated by the EA for ARM information are visualized and then the salient pieces of information (i.e. $<$feature,attribute$>$ pairs) are integrated into metro lines and subsequently into the whole story. From the story point of view, the $<$feature,attribute$>$ pairs of metro lines that are connected between each other with implication relations, represent the plot of the story. The sequence of relations also indicates the direction of the story plot. However, conflicts caused by the interrelation of metro stops of different metro lines are designated by vertical arcs. 

This study is focused on the visualization of the best result obtained by constructing the metro map of ARM information based on the five previously mentioned datasets. The best results after optimizing more instances obtained by constructing the metro lines by varying the metro path length $\tau\in[3,\text{APL}_G]$ and the number of metro lines $K\in[2,|\mathit{Source}(G)|]$ (Table~\ref{tab:experim}) are analyzed. 
\begin{table}[!htb]
\begin{center}
\small
\caption{Experimental setup by constructing the metro maps.}
\label{tab:experim}
\begin{tabular}{ l|r|r|c }
\hline
Dataset & Path length & Number of ML & Instances \\ \hline
Mushroom & $\tau\in[3,11]$ & $K\in[2,11]$ & 9 \\
Iris & $\tau\in[3,23]$ & $K\in[2,4]$ & 21 \\
Abalone & $\tau\in[3,5]$ & $K\in[2,10]$ & 3 \\
Wine & $\tau\in[3,8]$ & $K\in[2,8]$ & 6 \\
Sport & $\tau\in[3,6]$ & $K\in[2,5]$ & 4 \\
\hline
\end{tabular}
\normalsize
\end{center}
\end{table}

Thus, the best metro maps with the longer metro paths are compared several times with their counterparts with the best fitness, in order to highlight the interesting knowledge hidden in data. The results of the visualization of metro maps are illustrated in the remainder of the paper. Let us notice that all metro maps in the paper were visualized using the graphical tool embedded into the R programming environment.

\subsubsection{Mushroom dataset}
\begin{figure*}[!hbt]
\begin{minipage}{.65\linewidth}
\begin{minipage}{\linewidth}
    \centering
    \includegraphics[width=\linewidth]{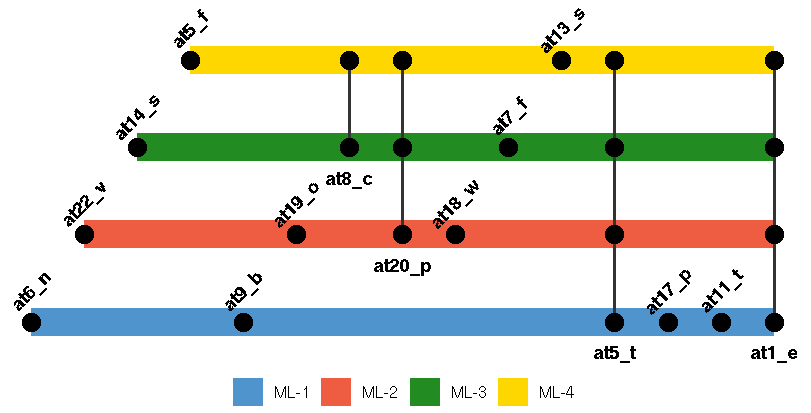}
    \caption{subfigure}{Path length $\tau=6$.}
    \label{fig:mush1}
\end{minipage}
\begin{minipage}{\linewidth}
    \centering
    \includegraphics[width=\linewidth]{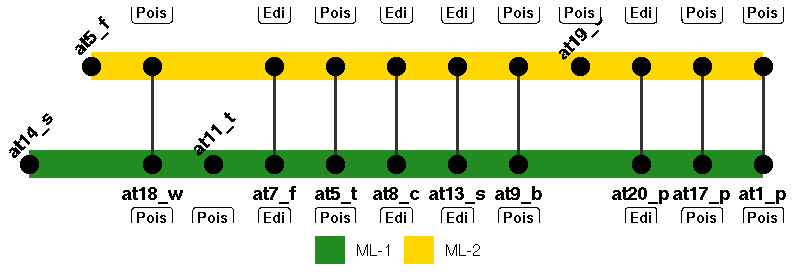}
    \caption{subfigure}{Path length $\tau=11$.}
    \label{fig:mush2}
\end{minipage}
\end{minipage}
\begin{minipage}{.3\linewidth}
 \centering
 \small
\begin{tabular}{ |c|l| }
\hline
Tag & Attribute \\	\hline
at1\_e & class\_edible \\
at1\_p & class\_poisonous \\
at5\_f & bruises?\_no \\
at5\_t & bruises?\_bruises \\
at6\_n & odor\_none \\
at7\_f & gill-attachment\_free \\
at8\_c & gill-spacing\_close \\
at9\_b & gill-size\_broad \\
at11\_t & stalk-shape\_tapering \\
at13\_s & stalk-surface-above-ring\_smooth \\
at14\_s & stalk-surface-below-ring\_smooth \\
at17\_p & veil-type\_partial \\
at18\_w & veil-color\_white \\
at19\_o & ring-number\_one \\
at20\_p & ring-type\_pendant \\
at22\_v & population\_several \\
\hline
\end{tabular}
\end{minipage}
\caption{Visualization of the Mushroom dataset.}
\label{fig:mushroom}
\end{figure*}

Visualization of the best run obtained by the construction of a metro map based on ARM information of the Mushroom dataset is presented in Fig.~\ref{fig:mushroom} that is divided into two diagrams: The upper depicts the metro map with metro lines (designated with ML-1 to ML-4) of path length $\tau=6$ (Fig.~\ref{fig:mushroom}a), and the lower the metro map with metro lines (designated with ML-1 and ML-2) of $\tau=11$ (Fig.~\ref{fig:mushroom}b). As a result, the first metro map consists of four, while the second one from two metro lines. Indeed, each metro map consists of metro stops designating the physical measurements identifying an edibility or poisonous of the observed mushrooms, where the searched characteristic is represented as a destination metro stop. Also, a legend is appended to the diagrams that transforms the tags denoting metro stops into $<$feature,attribute$>$ pairs.

Interestingly, all the metro lines in Fig.~\ref{fig:mushroom}a lead to the same destination (i.e., 'class\_edible' $<$feature/attribute$>$ pair), denoting the characteristics of the edible mushrooms. On the other hand, both the metro lines in Fig.~\ref{fig:mushroom}b arose already in the first metro map, but drove to the different destination designating the poisonous mushrooms. From this evolution from the edible to poisonous mushrooms, we can infer which ingredients cause that the edible mushroom becomes poisonous. In line with this, each metro stop (i.e., measurement) in this metro map is designated with the mark 'Edi', when the metro stop is already present in the corresponding metro line of the first metro map (e.g., 'gill-spacing\_close', 'ring-type\_pendant', 'gill-attachment\_free', 'stalk-surface-above-ring\_smooth', and 'bruises?\_bruises') and with mark 'Pois', when these are inherited from the other metro lines of the first metro map (i.e., 'veil-color\_white', 'stalk-shape\_tapering', and 'veil-type\_partial').  

In summary, although all metro stops arisen in the first metro lines led to the edible mushrooms, their mutual combination in the second metro map can determine whether they are edible or not. This means, there is no simple rule for determining the edibility of the mushroom. The mutual combination of metro stops can be inherited either from the same metro lines of the first metro map (like 'gill-attachment\_free' and 'stalk-surface-above-ring\_smooth') or the other metro lines (like 'gill-size\_broad', 'stalk-shape\_tapering', and 'veil-color\_white'). In the last case, the first two measures emerge as independent metro stops in the second metro map, while the third connects mutually to both the corresponding metro lines.

\subsubsection{Iris dataset}
\begin{figure*}[!hbt]
\begin{minipage}{.7\linewidth}
    \centering
    \includegraphics[width=\linewidth]{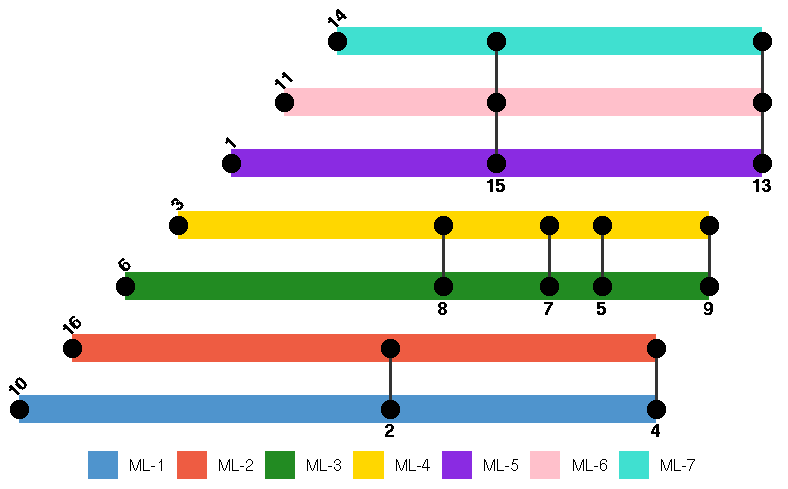}
\end{minipage}
\begin{minipage}{.3\linewidth}
 \centering
 \small
\begin{tabular}{ |c|l| }
\hline
Nr. & Attribute \\	\hline
1 & sepalwidth\_(2.6-3.2] \\ 
2 & petalwidth\_(0.7-1.3] \\
3 & sepallength\_(5.2-6.1] \\ 
4 & class\_Iris-versicolor \\ 
5 & sepallength\_(-inf-5.2] \\ 
6 & sepalwidth\_(3.2-3.8] \\ 
7 & petallength\_(-inf-2.47] \\ 
8 & petalwidth\_(-inf-0.7] \\ 
9 & class\_Iris-setosa \\ 
10 & sepalwidth\_(-inf-2.6] \\ 
11 & petalwidth\_(1.9-inf) \\
13 & class\_Iris-virginica \\
14 & sepallength\_(7-inf) \\
15 & petallength\_(5.42-inf) \\
16 & petallength\_(2.47-3.95] \\
\hline
\end{tabular}
\end{minipage}
\caption{Visualization of the Iris dataset (path length $\tau=5$).}
\label{fig:iris}
\end{figure*}

A visualization of the best results obtained by the construction of a metro map based on ARM information within the Iris dataset is presented in Fig.~\ref{fig:iris}, from which it can be seen that the corresponding metro map consists of seven metro lines of moderate path length ($\tau=5$). In this case, metro stops identify physical measurements of the observed iris plants, like sepal length/width and petal length/width, while the destination metro stop specifies their proper type.

Indeed, the first two metro lines (i.e., ML-1 and ML-2) started with sepal width less than 2.6~cm and petal length between 2.475 and 3.95~cm classifying the Iris versicolor class if the iris plant owns petal width between 0.7 and 1.3~cm. The next two metro lines (i.e., ML-3 and ML-4) describe how to recognize the setoza iris plant, starting with the floor of sepal width between 3.2 and 3.8~cm and the sepal length between 5.2 and 6.1~cm. These metro lines are interrelated, even with three metro stops (i.e., 'petalwidth\_(-inf-0.7]', 'petallength\_(-inf-2.475]', and 'sepallength\_(-inf-5.2]'). The last three metro lines (i.e., ML-5 to ML-7) indicate that the iris virginica has arisen from a plant with sepal width between 2.6 and 3.2~cm, or petal width more than 1.9~cm, or sepal length more than 7~cm. However, the petal length of these plants must be more than 5.425~cm.

Actually, the Iris dataset contains two clusters: (1) one containing 'Iris setosa', and (2) the other containing 'Iris virginica' and 'Iris versicolor'. Based on this assumption, the first two metro lines classifying the second cluster should be interrelated, but, surprisingly, this fact is not reflected in the mined metro map. Moreover, it turns out that when the species information is used as proposed by Fisher~\cite{fisher1936use}, all three classes are linearly separable~\cite{gorban2007topological}. Indeed, the advantage of the proposed algorithm for generating metro maps based on ARM information is that it already considers these clusters as linearly independent. 

\subsubsection{Abalone dataset}
\begin{figure*}[!hbt]
\begin{minipage}{.7\linewidth}
\begin{minipage}{\linewidth}
    \centering
    \includegraphics[width=\linewidth]{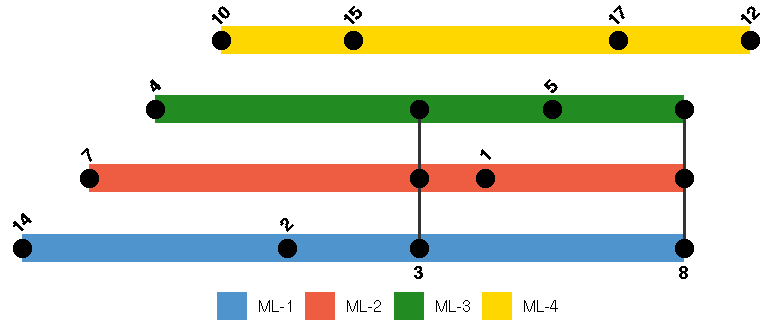}
    \caption{subfigure}{Path length $\tau=4$.}
    \label{fig:abalone1}
\end{minipage}
\begin{minipage}{\linewidth}
    \centering
    \includegraphics[width=\linewidth]{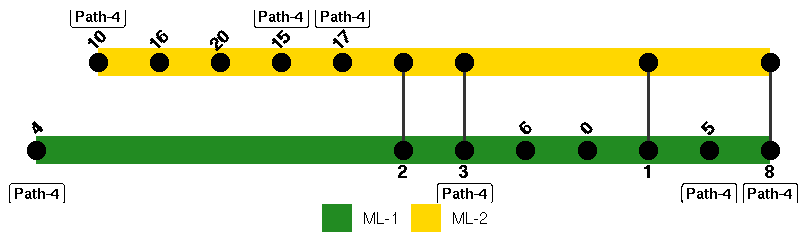}
    \caption{subfigure}{Path length $\tau=9$.}
    \label{fig:abalone2}
\end{minipage}
\end{minipage}
\begin{minipage}{.3\linewidth}
 \centering
 \small
\begin{tabular}{ |c|l| }
\hline
Nr. & Attribute \\	\hline
0 & shell\_(-inf-0.25] \\
1 & whole\_(-inf-0.71] \\
2 & height\_(-inf-0.28] \\
3 & diameter\_(-inf-0.20] \\
4 & length\_(-inf-0.26] \\
5 & viscera\_(-inf-0.19] \\
6 & shucked\_(-inf-0.37] \\
7 & sex\_I \\
8 & rings\_(-inf-8] \\
10 & shucked\_(0.74-1.12] \\
12 & rings\_(8-15] \\
14 & viscera\_(0.38-0.57] \\
15 & whole\_(1.41-2.12] \\
16 & length\_(0.63-inf) \\
17 & diameter\_(0.50-inf) \\
20 & shucked\_(0.37-0.74] \\
\hline
\end{tabular}
\end{minipage}
\caption{Visualization of the Abalone dataset.}
\label{fig:abalone}
\end{figure*}

A visualization of the best metro map obtained with the proposed EA by analysis of ARM information within the Abalone dataset are illustrated in Fig.~\ref{fig:abalone}. 
The figure is divided into two diagrams, where the former depicts a metro map with four metro lines (denoted as ML-1 to ML-4) and the latter with two metro maps (denoted as ML-1 and ML-2). As a result, the first metro map is constructed with path length $\tau=4$, and the second with path length $\tau=9$. The metro stops denote the physical measurements of the observed abalone (like sex, length, diameter etc.), and connect them into a path that leads towards the destination metro stop determining the number of rings. However, the age of the corresponding mushroom is determined on basis of this number.

Actually, the metro lines ML-3 and ML-4 in the first metro map are depicted as ML-1 and ML-2 in the second one, respectively. These measurements are denoted using mark 'Path-4' in the second metro map. Indeed, the metro line ML-4 in the first metro map determining the metro stops leads to the number of rings between 8 and 15 (i.e., age of 9.5 to 16.5) change direction in the second metro line of the second metro map from the same starting metro stop to the number of rings less than 8 (i.e., age of less than 9.5 years).

From comparison of both the metro maps, it can be concluded that there are two main reasons for changing the age of the abalone from between 9.5 and 16.5 years in ML-4 of the first metro map to the age of less than 9.5 years in the ML-2 of the second metro map: (1) adding two new $<$feature,attribute$>$ pairs into the ML-2 (i.e., the length of abalone higher than 0.63~mm, and the shucked weight between 0.37 and 0.74~grams), and (2) inheriting the $<$feature,attribute$>$ pairs from the ML-1 (i.e., height less than 0.28~mm and diameter less than 0.20~mm).  

\subsubsection{Wine dataset}
\begin{figure*}[!hbt]
\begin{minipage}{.7\linewidth}
\begin{minipage}{\linewidth}
    \centering
    \includegraphics[width=\linewidth]{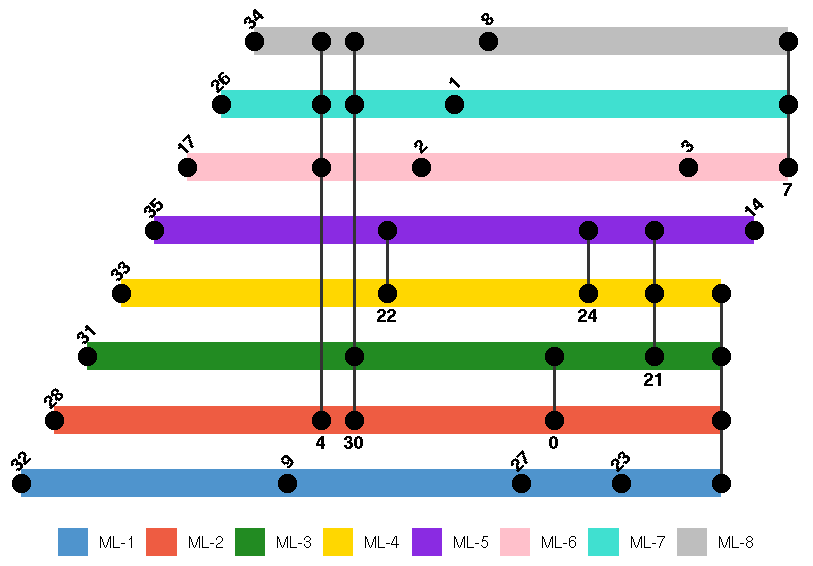}
    \caption{subfigure}{Path length $\tau=5$.}
    \label{fig:wine1}
\end{minipage}
\begin{minipage}{\linewidth}
    \centering
    \includegraphics[width=\linewidth]{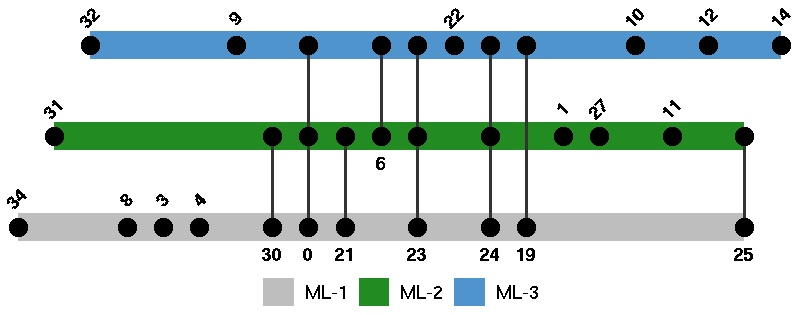}
    \caption{subfigure}{Path length $\tau=11$.}
    \label{fig:wine2}
\end{minipage}
\end{minipage}
\begin{minipage}{.3\linewidth}
 \centering
 \small
\begin{tabular}{ |c|l| }
\hline
Nr. & Attribute \\	\hline
0	&	alcalinity\_(15.45-20.3]	\\
1	&	ash\_(2.29-2.76]	\\
2	&	malic\_acid\_(-inf-2.00]	\\
3	&	alcohol\_(12.93-13.88]	\\
4	&	flavanoids\_(2.71-3.89]	\\
6	&	C	\\
7	&	class\_1	\\
8	&	color\_intensity\_(4.21-7.14]	\\
9	&	flavanoids\_(1.52-2.71]	\\
10	&	hue\_(0.79-1.09]	\\
11	&	magnesium\_(-inf-93]	\\
12	&	color\_intensity\_(-inf-4.21]	\\
14	&	class\_2	\\
17	&	hue\_(1.09-1.40]	\\
19	&	magnesium\_(93-116]	\\
20	&	total\_phenols\_(1.70-2.43]	\\
21	&	total\_phenols\_(-inf-1.70]	\\
22	&	flavanoids\_(-inf-1.52]	\\
23	&	hue\_(-inf-0.79]	\\
24	&	proanthocyanins(-inf-1.20]	\\
25	&	class\_3	\\
26	&	color\_intensity\_(7.14-10.07]	\\
27	&	nonflavanoid\_phenols\_(0.39-0.53]	\\
28	&	nonflavanoid\_phenols\_(-inf-0.26]	\\
29	&	malic\_acid\_(2.00-3.27]	\\
30	&	proanthocyanins(1.99-2.79]	\\
31	&	alcohol\_(13.88-inf)	\\
32	&	alcohol\_(-inf-11.98]	\\
33	&	malic\_acid\_(3.27-4.53]	\\
34	&	magnesium\_(116-139]	\\
35	&	nonflavanoid\_phenols\_(0.53-inf)	\\
\hline
\end{tabular}
\label{tab:iris}
\end{minipage}
\caption{Visualization of the Wine dataset.}
\label{fig:wine}
\end{figure*}

Visualizing the metro map that emerged on the basis of ARM information within the Wine dataset is depicted in Fig.~\ref{fig:wine}, that is divided into two diagrams, representing the metro maps constructed by path lengths $\tau=5$ and $\tau=11$. 
Thereby,  metro stops designate the quantities of chemical constituents found by analyzing, where each destination metro stop determines one of the three types of wines. 

Indeed, the first metro map consists of seven metro lines, while the second of only three. The metro lines of the same colors emerging in both metro maps, in some way, reproduce the development of the three different wine types which originated  from the same source during the time. The older the wine, the less complex is its chemical structure. This means that once simple chemical structure found in the analysis of the older wine types became more and more complex over time. Interestingly, this trend can be observed in the constructed metro maps.

For instance, the ML-7, ML-3, and ML-1 in the first metro map emerged as ML-1, ML-2, and ML-3 in the second metro map, respectively. Thus, the first four metro lines classify quantities of constituents needed for producing the 'class-3', ML-5 for the 'class-2' and the remaining three metro lines for 'class-1' type of wine. The second metro map incorporates only constituents found in the 'class-3' and 'class-2' types of wine. 

Comparison between both metro maps revealed, in which direction the three different cultivars proceeded. Thus, the ML-1 in the second metro map originally describes the chemical analysis needed by producing the 'class-1' type of wine. Although all original chemical constituents, characteristic for 'class\_1' can also be found in the new metro map (i.e., color\_intensity\_(4.21-7.14], alcohol\_(12.93-13.88], flavanoids\_(2.71-3.89], proanthocyanins(1.99-2.79]), the constituents determining the 'class\_3' wine type were inherited from the first three metro lines of the first metro map (i.e., alcalinity\_(15.45-20.3], malic\_acid\_(-inf-2.00], hue\_(-inf-0.79], proanthocyanins(-inf-1.20]), as well as new ones have emerged (as 19). The ML-2 in the second metro map remains unchanged regarding the destination metro stop and also inherits all three constituents from the ML-3 of the original metro map. Besides the interrelation metro stops with the ML-1, there also emerged two new constituents determining the same class (i.e., proanthocyanins(-inf-1.20], magnesium\_(-inf-93]). The ML-3 suffers much of the modifications regarding ML-1 in the original metro map. At first, the destination of the metro map changes from the original 'class\_2' to the 'class\_2'. Then, two constituents are inherited from the original 'class\_2' (i.e., flavanoids\_(1.52-2.71], flavanoids\_(-inf-1.52]), while the next four shared with the new 'class\_3' (i.e., alcalinity\_(15.45-20.3], hue\_(-inf-0.79], proanthocyanins(-inf-1.20], magnesium\_(93-116]). Finally, hue\_(0.79-1.09] and color\_intensity\_(-inf-4.21] are emerged anew. 

In summary, the 'class\_1' disappears from the second metro map, but some constituents were inherited from it and emerged in the new 'class\_3'. On the other hand, the 'class\_2' type of wine stemmed from the 'class\_3' appearing in this metro map. However, some new constituents were added to this.
 
\subsubsection{Sport dataset}
\begin{figure*}[!htb]
    \centering
    \includegraphics[width=0.7\linewidth]{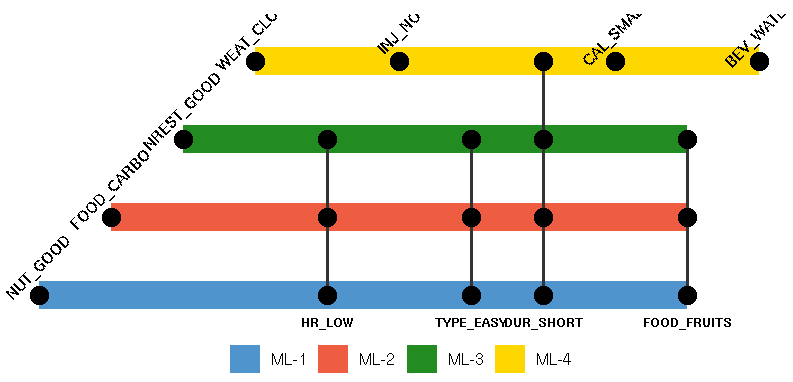}
    \caption{Visualization of the Sport dataset (path length $\tau=5$).}
    \label{fig:Sport}
\end{figure*}

Visualization of the best metro map obtained on the basis of ARM information hidden within the Sport dataset is presented in Fig.~\ref{fig:Sport}, from which it can be seen that this map consists of four metro lines of length $\tau=5$. Each metro line describes paths from the source to one of five destination metro stops, where the particular metro stop denotes the particular psycho-physical measurement obtained by the observed athlete during a training session. 

The metro map is strongly interrelated, due to several connected metro stops. For instance, the first three metro lines ML-1 to ML-3 describe paths from different starting metro stops toward the same destination 'FOOD\_FRUITS', reached by even the same interrelated intermediate metro stops. The fourth metro line, on the other hand, leads to the destination metro stop 'BEV\_WATER' from the 'WEAT\_CLOUD' starting metro stop. Thus, all four metro lines are interrelated by the 'DUR\_SHORT' metro stop. 

The knowledge hidden in the presented metro map complies with the theory of sport, and is typical for an amateur athlete overcoming the easier stress during training. Actually, the metro map describes the food and beverage needs of an athlete during the realization of sport training sessions of low intensity. In line with this, it holds that the lower calorie food is necessary for easier training of short duration. Additionally, the fitter athlete with no injuries does not need any isotonic beverage by realizing the training sessions of short duration in cloudy weather.

\subsection{Discussion}
A metro map of information is a new visualization method that has taken inspiration from real metro maps. Metaphorically, like real metro maps help people understand their surroundings, the metro maps help them understand their information landscape. Moreover, the structure of metro maps is designed so that these can even tell stories to users. Indeed, metro maps consist of metro lines that present different aspects of the same story. The metro lines contain metro stops that represent salient pieces of information. When these pieces interrelate between different metro lines, conflicts arise. These conflicts lead the plot of the story towards its resolution (the final metro stop). On the other hand, each story must have its own introduction (the destination metro stop). 

Let us emphasize that the metro map of ARM information presents a generalization of all the association rules in the transaction database and, therefore, ignores the special cases. Consequently, the extracted association rule sequences in metro lines only highlight the most general truths. As a result, it can happen that some conclusion does not hold true for each cases using this method. Despite everything, the method offers a new aspect in the visualization of ARM rules, and can help users direct their attention to the more salient pieces of information.

The core of information cartography in ARM presents the construction of a metro map, which is defined as an optimization problem that has been solved by an EA. The task of the EA is to find the best paths (representing metro lines) within the attribute graph according to predefined objectives that define a structure of the designed metro map. 
The characteristics of the constructed metro maps are as follows: The shorter metro lines pass the basic truths representing
building blocks, when the path lengths become longer. Actually, the building blocks connect together in longer metro lines and represent a complex knowledge hidden inside metro maps. On the other hand, the history of emerging some constituents, quantities of chemical substrates, or psycho-physical measurements can be predicted by evolution from shorter metro lines to the longer. Obviously, this evolution represents a potential idea to form the story with its introduction, plot of the play and conflicts that lead to the final resolution. Indeed, the story tells users even more than many other standard visualizations which can be revealed by our wide experimental work.

Let us see what some of observed metro maps narrate. For instance, the mushroom metro map could help the screenwriter to select the mushrooms with those constituents that are faithful for a character that must die according the screenplay due to poisoning. Abalone is known in cooking as a reach nutritional food resource that is full of high proteins, and low in fat and cholesterol. It represents a source of omega-3 fatty acids having a low risk of heart disease. The economic value of abalone is correlated positively with age~\cite{hossain2019econometric}. Especially interesting for economy are so-called 'cocktail' abalones whose shell weights generally measure 7-11~cm. In the sense of a restaurant owner serving the healthy dishes, among which are abalones in the first place, it is very important to determine the age of the abalones being sold on the local fish market without expensive experiments by counting the number of rings under the microscope. In line with this, the abalone metro map informs us that physical measurements characteristic for older abalones (like a diameter of more than 0.5~mm and whole weight between 1.41 and 2.12~grams) does not mean that the observed abalone is not the 'cocktail' one. 

However, as the stories describe specific situations, our metro maps of ARM information also stem from generalized extracted information that, in specific situations, sometimes does not hold up entirely. Despite this weakness, the technology of information cartography in ARM shows that there are new aspects in extracting the structured knowledge hidden in data and, especially, in transferring this knowledge to the user. The Wine story can undoubtedly be considered in this class. 

The Wine story is one of the most fascinating, while the point of the story confirms the power of the selected visualization. Without any knowledge about the real situation that happened in the specific region in Italy, our story, inferred from the corresponding metro map, speculates that there was an original wine type, let's say 'class\_1', which passes its good characteristics to its offspring plants in the same cultivar, let's say 'class\_3'. This type of wine has been evolved, and by adding the new chemical constituents caused an emerging of the new type of wine, let's say 'class-2'. Finally, we got three types of plants steam from the same ancestor that have been evolved in three different directions, although they grow in the same area under the same conditions.

In summary, the goal of the visualization process using metro maps is not only to present information to the user in an understandable way, but to assemble the constituent salient information pieces of the metro map into the whole. Moreover, this technology is capable of directing the user's decision-making process and simulating exactly what consequences false decisions can have.

\section{Conclusion}
\label{conclusion}

Nowadays, we are confronted with the large-scale creation of unstructured data that are hard to analyze manually. In line with this, a lot of ML methods have arisen, with the purpose of discovering new information hidden within data. One of these methods is also ARM, devoted to discovering the interesting relations between attributes in huge transaction databases. Normally, the algorithms for ARM generate a huge number of association rules collected in datasets in an unstructured form. From these datasets it is not so easy to extract structured knowledge and present this in a form that is automatically appropriate for ordinary users.

This paper proposes a new method for creating metro maps of ARM information automatically (also information cartography) that consist of five steps: creating an ARM dataset, association rule simplification, attribute graph definition, metro map construction, and metro map visualization. Actually, the contribution belongs to the XAI domain that has recently been achieving a notable momentum with the progress of deep learning. As a result, the study revealed that the ARM information cartography is suitable for explaining knowledge hidden in ARM databases on the one hand, and the concept spreads the applicability of the information cartography to the other ML methods on the other. Moreover, the EA for constructing the metro maps was applied primarily to the information cartography, where no, or a less domain specific knowledge, exists. The results on five ARM databases showed that the constructed metro maps are robust enough for using for explanation purposes in practice.

In the future, information cartography could also be applied to the other UCI Machine Learning datasets. Particularly, its application to the sports domain could be very interesting, where this technology could be used for Interactive Machine Learning (iML) and, thus, help sport athletes optimize their learning behavior during a sport training session through interaction with the proposed information cartography.

\ifCLASSOPTIONcompsoc
  \section*{Acknowledgments}
\else
  \section*{Acknowledgment}
\fi

Iztok Fister Jr. is grateful for the financial support from the Slovenian Research Agency (Research Core Funding No. P2-0057). Iztok Fister is grateful the financial support from the Slovenian Research Agency (Research Core Funding No. P2-0042 - Digital twin).

\end{document}